\documentclass[runningheads]{llncs}

\usepackage[T1]{fontenc}
\usepackage{graphicx,amsmath,amssymb}
\usepackage{algorithm}
\usepackage{algorithmic}

\usepackage{makecell}
\usepackage{multirow}
\usepackage{microtype}
\usepackage{cuted}
\usepackage{soul}

\usepackage{mwe} % to get dummy images
\usepackage{ragged2e}

\usepackage{color}

\begin{document}
\title{FedDermaSeg: Federated Learning for Dermatological Image Segmentation}

\author{Anabik Pal\inst{1} \and
Ganesh Patidar\inst{2} \and
Bikash Santra\inst{3}}

\authorrunning{A. Pal et al.}

\institute{Department of Computer Science, Indian Institute of Science Education and Research Berhampur, India \and
Department of Mathematics, Indian Institute of Technology Jodhpur, India \and School of Artificial Intelligence and Data Science, Indian Institute of Technology Jodhpur, India\\
\email{Corresponding author: anabikpal@iiserbpr.ac.in}\\
}

\maketitle              % typeset the header of the contribution
\begin{abstract}
Skin cancer is a global health concern. Early detection of skin cancer lesions and treatment planning are important for reducing premature deaths. To deal with the dermatologists' limitations, automated systems are highly demanding for examining skin conditions. Lesion segmentation is one of the prerequisites for developing any computerised system. The commonly used deep segmentation networks are trained in a centralized manner, i.e. images, and their annotated segmentation masks are gathered in a centralized server, and the training is performed. This centralized training loses data privacy, which is an important concern for the medical domain. Moreover, a huge computation has to be conducted on a single server. Hence, the feasibility of applying federated learning is experimented with and analysed for skin segmentation model development. The train and valid set of the ISIC 2018 skin lesion segmentation challenge dataset is used to simulate the distributed learning environment and build a federated model. The test set of the ISIC 2018 skin lesion segmentation challenge dataset and $PH^2$ dataset are used for performance evaluation. As per experimental results, the federated model produces comparable performance with the centralized training and improves the local models.

\end{abstract}

\section{Introduction}

Skin cancer is a major threat to our society. The early detection can reduce premature death. The development of an automated system is required to reduce the experts' workload and enhance medicare for early detection of such skin conditions. Automated skin image segmentation algorithm development is an important pre-requisite for such automated systems~\cite{Tschandl2018}. Although several classical image processing techniques are proposed~\cite{Barata2019}, their performance is limited and prone to misclassification of the image pixels. In contrast, the application of the latest deep learning-based approaches will be a better choice as they have proven to be efficient in many image analysis applications~\cite{Kshatri2023}. Broadly, the segmentation network can be built with deep convolutional neural network (DCNN) based encoder-decoder-like architecture~\cite{Ronneberger2015u} or transformer-based architecture~\cite{Carion2020}. In deep networks of the first kind, the encoder obtains the low-dimensional feature representation of the input images, and the decoder finds the semantic maps of the pixels using the low-dimensional feature representation. On the other hand, transformer-based approaches find low-dimensional image representation with attention mechanisms, and by combining the feature representation with object queries, the object regions are obtained. Note that the DCNN-based architectures fail to capture the global context of an image during feature learning~\cite{YANAGISAWA2023}, and the transformer networks are not capturing local interactions in the feature maps~\cite{Xin2024}. Hence, several segmentation architectures are designed that hybridize the potential of both CNN and transformer networks~\cite{CHEN2024} and applied in skin lesion segmentation purposes~\cite{DING2024,HUANG2024}. 
%%%%%%%%%%%%%

The key source of high performance in the above-mentioned approaches lies in the use of a huge volume of data samples during training. Getting a good volume of curated and annotated dataset(s) from a single clinic or hospital is often challenging. The easiest solution is to use a single repository and gather data prepared by all collaborated sources. However, this may violate data privacy and clinical ethics. Hence, federated learning is used as an effective alternative to train a robust model in collaboration with multiple sources without violating data privacy. In spite of data privacy control, federated learning distributes the computation among the collaborated parties (sources) and makes the computation de-centralized~\cite{Yang2019}. In general setting, federated learning is performed with the help of a parameter server. First, the parameter server creates a deep model and initializes the weight matrices. Then the model is sent to the collaborated clients for weight update based on their locally available dataset. Then, the updated models are transferred to the server for aggregation, which produces a global collaborative model. This process continues until the model converges. The model aggregation is performed with the FedAvg algorithm, which uses a weighted average of the weight matrices~\cite{McMahan2016}.

This paper develops a deep segmentation model with federated learning with FedAvg algorithm~\cite{McMahan2016} in the cross-silo setting, systematically evaluating skin lesions' performance and analyzing its performance trade-offs. Equal weights are given to the clients during model aggregation in the server. As we are interested in analysing the algorithmic effectiveness of the proposed approach, we create a synthetic environment to create the clients and the parameter server. ISIC 2018 skin lesion segmentation challenge training dataset~\cite{codella2019skin} is used for training the federated segmentation model. After training, the performance of the federated model is analysed with ISIC 2018 skin lesion segmentation challenge test dataset and $PH^2$ dataset~\cite{mendoncca2015ph2}. Literature shows that some works are available to develop deep skin image classification models~\cite{Bdair2021} with federated learning. However, as per our survey, limited work attempts to develop a federated segmentation model for skin lesion segmentation. In contrast to that, this work employs a federated learning algorithm for building a skin lesion segmentation model, analyzes the performance of commonly used deep segmentation networks and analyzes the performance of commonly used segmentation loss functions.

The rest part of the paper is organized as follows: Section~\ref {sec:Methodology} discusses the proposed methodology; Section~\ref{sec:Experiments} presents the experimental setup; the experimental analysis is kept at Section~\ref{sec:Results}; and finally, Section~\ref{sec:Conclusion} concludes the paper.

\section{Methodology}\label{sec:Methodology}

\begin{figure*}[!t]
    \centering
      \includegraphics[width=.99\textwidth]{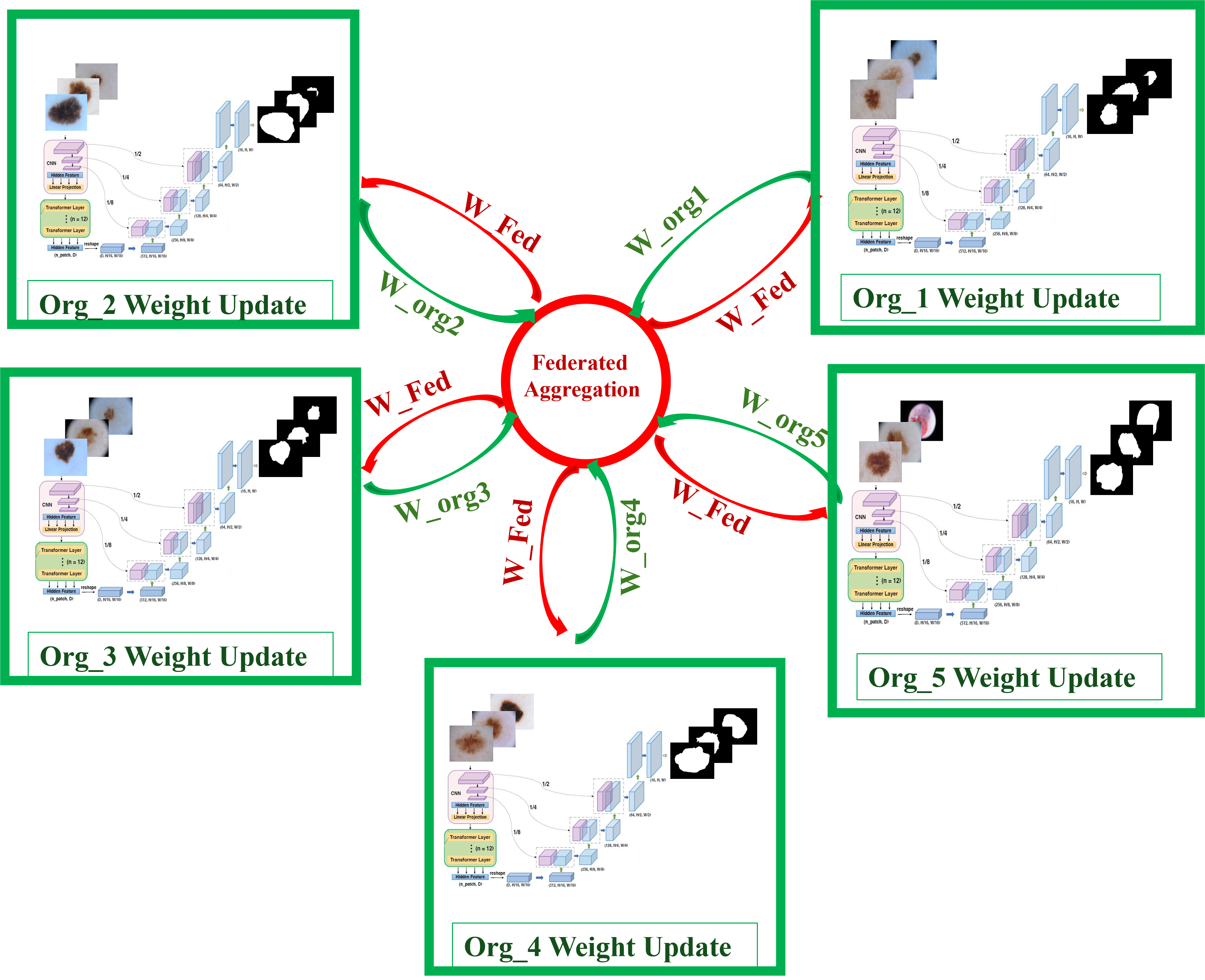} % Adjust file name if needed
    \caption{The flow diagram of training segmentation network with federated learning. $Org_i$ refers to $i^{th}$ client; $W\_Orgi$ refers to local weight update in $i^{th}$ client during a communication round; $W\_Fed$ refers to global model after local model aggregation.}
   \label{fig:flow_diagram}
\end{figure*}

The block diagram of our work is presented in Fig~\ref{fig:flow_diagram}. Fig~\ref{fig:flow_diagram} shows that there are five clients surrounded by green borders where local model training is happening. Note that in this paper, we are considering five clients; however, a finite number of clients can participate in the federated training process. The parameter server is shown in the middle part of this diagram and shown as a red-marked circle. At the beginning, a deep image segmentation network with initial weight matrices is sent to all clients. On receiving the model and its parameters, each client trains the model with their local annotated images. After $E$ number of local model updates, the updated models are sent to the server for aggregation. The server then aggregates the clients' models to have a single model. This process is continued for $T$ rounds of communications. Note that we apply the federated average algorithm~\cite{McMahan2016} for preparing the global model and consider equal weights for all clients. For the reader's convenience, the training mechanism is given in the Algorithm~\ref{alg:federated_learning}.  

\begin{algorithm}
\caption{Federated Learning Algorithm}
\label{alg:federated_learning}
\begin{algorithmic}[1]
\REQUIRE $\{D_i\}_{i=1}^N$: datasets for $N$ clients; $T$: total rounds; $E$: local epochs; $\eta$: learning rate  
\ENSURE Global model $M_G$
\STATE Initialize global model $M_G$ with random weights $\theta_G$
\FOR{$t = 1$ \TO $T$}
    \STATE Broadcast $\theta_G$ to all clients
    \FORALL{clients $i \in \{1, \dots, N\}$ (in parallel)}
        \STATE Initialize $M_i \gets \theta_G$
        \FOR{$e = 1$ \TO $E$}
            \STATE Train $M_i$ on $D_i$ with learning rate $\eta$ to get $\theta_i$
        \ENDFOR
    \ENDFOR
    \STATE Aggregate: $\theta_G \gets \frac{1}{N} \sum_{i=1}^N \theta_i$
\ENDFOR
\STATE \RETURN $M_G$
\end{algorithmic}
\end{algorithm}

Regarding segmentation architecture, we experiment with two state-of-the-art deep image segmentation networks, namely UNet~\cite{Ronneberger2015u} and TransUNet~\cite{CHEN2024}. In short, the UNet architecture contains a series of convolution layers and deconvolution layers. The role of convolutional layers is to find the lower-dimensional feature representation of the input images, and the deconvolutional layers are kept to obtain the segmentation masks in the original dimension. In UNet, there is neuronal interaction in the local regions only. In contrast, TransUNet uses Convolutional layers to find the lower-dimensional representation of the input image with local neuronal interaction. Then, the lower-dimensional representations are processed in the transformer layers. In the transformer layers, the final image embedding is prepared with global neuronal interactions. The learnt feature is then processed with deconvolutional layers. Both of the networks are trained with the above-mentioned federated learning algorithms and compared.

As per our workflow, the training or weight update happens only for the clients. We use a combination of the following state-of-the-art loss functions: (i) Binary Cross Entropy (BCE), (ii) Dice Loss, (iii) Focal Loss. Note that the lesional areas are much less than the normal skin regions in the input images; hence, the BCE loss can be biased towards detecting normal regions. This motivate us to incorporate Dice Loss and Focal Loss, which have some capability to deal with this bias. The mathematical definition of those loss functions are given below:

\begin{equation}
\text{BCE}(y, \hat{y}) = -\left( y \cdot \log(\hat{y}) + (1 - y) \cdot \log(1 - \hat{y}) \right)
\end{equation}

\begin{equation}
\text{Dice Loss}(y, \hat{y}) = 1 - \frac{2 \sum_{i=1}^{N} y_i \hat{y}_i}{\sum_{i=1}^{N} y_i + \sum_{i=1}^{N} \hat{y}_i + \epsilon}
\end{equation}

\begin{equation}
\text{Focal Loss}(y, \hat{y}) = - \alpha \left(1 - \hat{y}\right)^{\gamma} y \cdot \log(\hat{y}) - (1 - \alpha) \, \hat{y}^{\gamma} (1 - y) \cdot \log(1 - \hat{y})
\end{equation}

In the above equations, $y$ is the ground-truth mask; $\hat{y}$ is the predicted mask; $N$ is the total number of pixels; $\alpha, \gamma$ are scalar parameters; $\epsilon$ is a small number which is dealing with division by zero error.

\section{Experimental Setup}\label{sec:Experiments}
\subsection{Datasets}
We use the following two publicly available dermoscopic image datasets for our research:\\
\noindent
\textbf{ISIC:}
The International Skin Imaging Collaboration (ISIC) prepared multiple dermoscopy image datasets and made them publicly available as challenge datasets~\cite{codella2019skin}. We use the \textbf{ISIC 2018 skin image segmentation challenge} dataset as it contains the maximum number of skin lesion segmentation ground-truths. The dataset is organised into three subsets: training ($2,594$), validation ($100$), and test ($1,000$). We use available lesion segmentation ground truth annotations for our research. Note that the images have varying appearances and contain diverse clinical conditions, such as melanoma, melanocytic nevus, basal cell carcinoma, squamous cell carcinoma, vascular lesions, dermatofibroma, and benign keratosis.

\justify
\textbf{PH$^2$:}
PH$^2$ dataset was made publicly available in 2013 by the Universidade do Porto, Instituto Superior Técnico Lisboa, and the Hospital Pedro Hispano Dermatology Service~\cite{mendoncca2015ph2}. It contains $200$ annotated (lesion ground-truth) dermoscopy images having the following three categories: nevus ($80$), atypical ($80$), and melanomas ($40$).

\subsection{Data Simulation for FL Setup}\label{sect:Data Simulation}
This study simulates the decentralized learning environment. We consider five clients so we need to have five disjoint datasets that will serve as local datasets. For that, we split the \textbf{training set} of the ISIC 2018 skin lesion segmentation challenge dataset into five disjoint sets- four sets containing $519$ images and the remaining one containing $518$ images. We hypothesize that simulating datasets through random splitting introduces a certain degree of non-IID data distribution across clients, mirroring what is commonly observed in real-world distributed learning scenarios. The \textbf{validation set} of the ISIC 2018 skin lesion segmentation challenge is used to assess the performance of the local training updates as well as the global model updates after federated aggregation. Both the \textbf{test set} of the ISIC 2018 skin lesion segmentation challenge and the \textbf{$PH^2$ dataset} are used for the performance evaluation of the trained model.

%%%%%%%%%%%%%%%%%%%%%%
\begin{table}[!t]
\renewcommand{\arraystretch}{1.2}
\centering
%\caption{Results of various models trained and tested with the ISIC 2018 dataset} 
\caption{Performance in the test set of ISIC 2018 skin lesion segmentation challenge dataset} 
% \vspace{0.1in}
\label{tab:model_results}
\scalebox{0.99}{
\begin{tabular}{llcccccccc}
\hline
\textbf{Network} & \textbf{Model} & \textbf{Dice} & \textbf{IoU} & \textbf{Accuracy} & \textbf{Precision} & \textbf{Recall} & \textbf{F1}  \\ \hline
\multirow{6}{*}{UNet} & local-1 & 0.882 & 0.789 & 0.937 & 0.934 & 0.835 & 0.882 \\ \cline{2-8}
& local-2 & 0.883 & 0.791 & 0.936 & 0.92 & 0.849 & 0.883 \\ \cline{2-8}
& local-3 & 0.880 & 0.786 & 0.935 & 0.912 & 0.850 & 0.880 \\ \cline{2-8}
& local-4 & 0.877 & 0.782 & 0.933 & 0.909 & 0.848 & 0.877 \\ \cline{2-8}
& local-5 & 0.881 & 0.787 & 0.935 & 0.911 & 0.853 & 0.881 \\ \cline{2-8}
& global-L & 0.919 & 0.850 & 0.950 & 0.908 & 0.930 & 0.919 \\ \cline{2-8}
& global-FL & 0.899 & 0.816 & 0.945 & 0.934 & 0.866  & 0.899 \\ \hline\hline
\multirow{6}{*}{TransUNet} & local-1 & 0.885 & 0.793 & 0.936 & 0.901 & 0.869 & 0.885 \\ \cline{2-8}
 & local-2 & 0.883 & 0.791 & 0.936 & 0.905 & 0.862 & 0.883 \\ \cline{2-8}
 & local-3 & 0.888 & 0.799 & 0.938 & 0.904 & 0.873 & 0.888 \\ \cline{2-8}
 & local-4 & 0.878 & 0.783 & 0.934 & 0.923 & 0.838 & 0.878 \\ \cline{2-8}
 & local-5 & 0.878 & 0.783 & 0.934 & 0.923 & 0.838 & 0.878 \\ \cline{2-8}
 & global-L & 0.927 & 0.864 & 0.959 & 0.933 & 0.921 & 0.927 \\ \cline{2-8}
 & global-FL & 0.905 & 0.828 & 0.946 & 0.904 & 0.907  & 0.905 \\ \hline
\end{tabular}}
\end{table}

\subsection{Implementation Details and Performance Metrics}
\label{sec:typestyle}
The PyTorch deep learning framework~\cite{Paszke2019} is used for implementing the proposed algorithm. The following deep segmentation networks: UNet~\cite{Ronneberger2015u} and TransUNet \cite{HUANG2024} are considered for experimental evaluation. For federated learning, the local models update weights for $K$ interaction numbers before aggregating in the central server. The whole weight update is performed for $T$ rounds. The value of $K$, $T$ is decided experimentally as presented in Section~\ref{sec:Results}. The models are trained with a stochastic gradient descent algorithm with an initial learning rate of $0.001$ and momentum of $0.9$. An experiment is carried out to determine the right combination of losses in our ablation study presented in Section~\ref{sec:Results}. For comparison with the local models built with the local dataset and the centralized training having the whole dataset, an initial learning rate of $0.001$ and momentum of $0.9$ is used.

%\subsection{Quantitative Evaluation}
In this work, the skin lesion segmentation models are evaluated with the following five commonly used metrics for performance analysis: Dice similarity coefficient (Dice), intersection-over-union (IoU) also known as Jaccard index or Jaccard similarity coefficient, accuracy, precision, recall, and F1 score \cite{muller2022towards}.

\section{Results and Discussions}\label{sec:Results}
%\label{sec:majhead}
This section discusses the experimental results obtained from our research. We are interested in comparing the performances of the federated learnt model with the following two scenarios: (i) the local models with their own data and (ii) a single model prepared with centralized training having all images and their ground truths. The first competing approach will advocate the efficacy of the federally learnt model, and the second one can provide an idea of how much the performance is dropped when centralised training is impossible. The performance with respect to the test set of ISIC 2018 skin lesion segmentation is available at Table~\ref{tab:model_results} and the performance with the images available in $PH^2$ dataset is available in Table~\ref{tab:TransUNet_results_PH2}. In the rest of this section, the models trained locally with its own dataset (as mentioned in Section~\ref{sect:Data Simulation}) are denoted by \textbf{`local-1'}, \textbf{`local-2'}, \textbf{`local-3'},\textbf{`local-4'}, and \textbf{`local-5'}; the model trained with federated learning is denoted as \textbf{`global-FL'}; finally, the models trained centrally with the full dataset is denoted by \textbf{`global-L'}.
\begin{table}[!t]
\renewcommand{\arraystretch}{1.2}
\centering
\caption{Results of various models trained with ISIC 2018 and tested with the $PH^2$ dataset} 
% \vspace{0.05in}
\label{tab:TransUNet_results_PH2}
\scalebox{0.99}{
\begin{tabular}{llcccccccc}
\hline
\textbf{Network} & \textbf{Model} & \textbf{Dice} & \textbf{IoU} & \textbf{Accuracy} & \textbf{Precision} & \textbf{Recall} & \textbf{F1}  \\ \hline
\multirow{6}{*}{UNet} & local-1 & 0.854 & 0.745 & 0.911 & 0.905 & 0.808 & 0.854 \\ \cline{2-8}
& local-2 & 0.851 & 0.741 & 0.907 & 0.88 & 0.824 & 0.851 \\ \cline{2-8}
& local-3 & 0.858 & 0.751 & 0.911 & 0.887 & 0.831 & 0.858 \\ \cline{2-8}
& local-4 & 0.847 & 0.735 & 0.903 & 0.861 & 0.833 & 0.847 \\ \cline{2-8}
& local-5 & 0.846 & 0.733 & 0.905 & 0.888 & 0.808 & 0.846 \\ \cline{2-8}
& global-L & 0.869 & 0.769 & 0.915 & 0.862 & 0.877 & 0.869 \\ \cline{2-8}
& global-FL & 0.877 & 0.781 & 0.922 & 0.893 & 0.861  & 0.877 \\ \hline\hline

\multirow{6}{*}{TransUNet} & local-1 & 0.886 & 0.796 & 0.927 & 0.886 & 0.887 & 0.886 \\ \cline{2-8}
 & local-2 & 0.847 & 0.735 & 0.899 & 0.83 & 0.866 & 0.847 \\ \cline{2-8}
 & local-3 & 0.879 & 0.784 & 0.921 & 0.875 & 0.882 & 0.879 \\ \cline{2-8}
 & local-4 & 0.883 & 0.791 & 0.928 & 0.928 & 0.842 & 0.883 \\ \cline{2-8}
 & local-5 & 0.885 & 0.794 & 0.926 & 0.892 & 0.878 & 0.885 \\ \cline{2-8}
 & global-L & 0.894 & 0.809 & 0.930 & 0.872 & 0.918 & 0.894 \\ \cline{2-8}
 & global-FL & 0.885 & 0.793 & 0.924 & 0.872 & 0.898  & 0.885 \\ \hline

\end{tabular}}
\end{table}

%%%%%%%%%%%%%%%%%%
\begin{figure}[!b]
    \centering
    % First image at the left
    \includegraphics[width=0.995\textwidth]{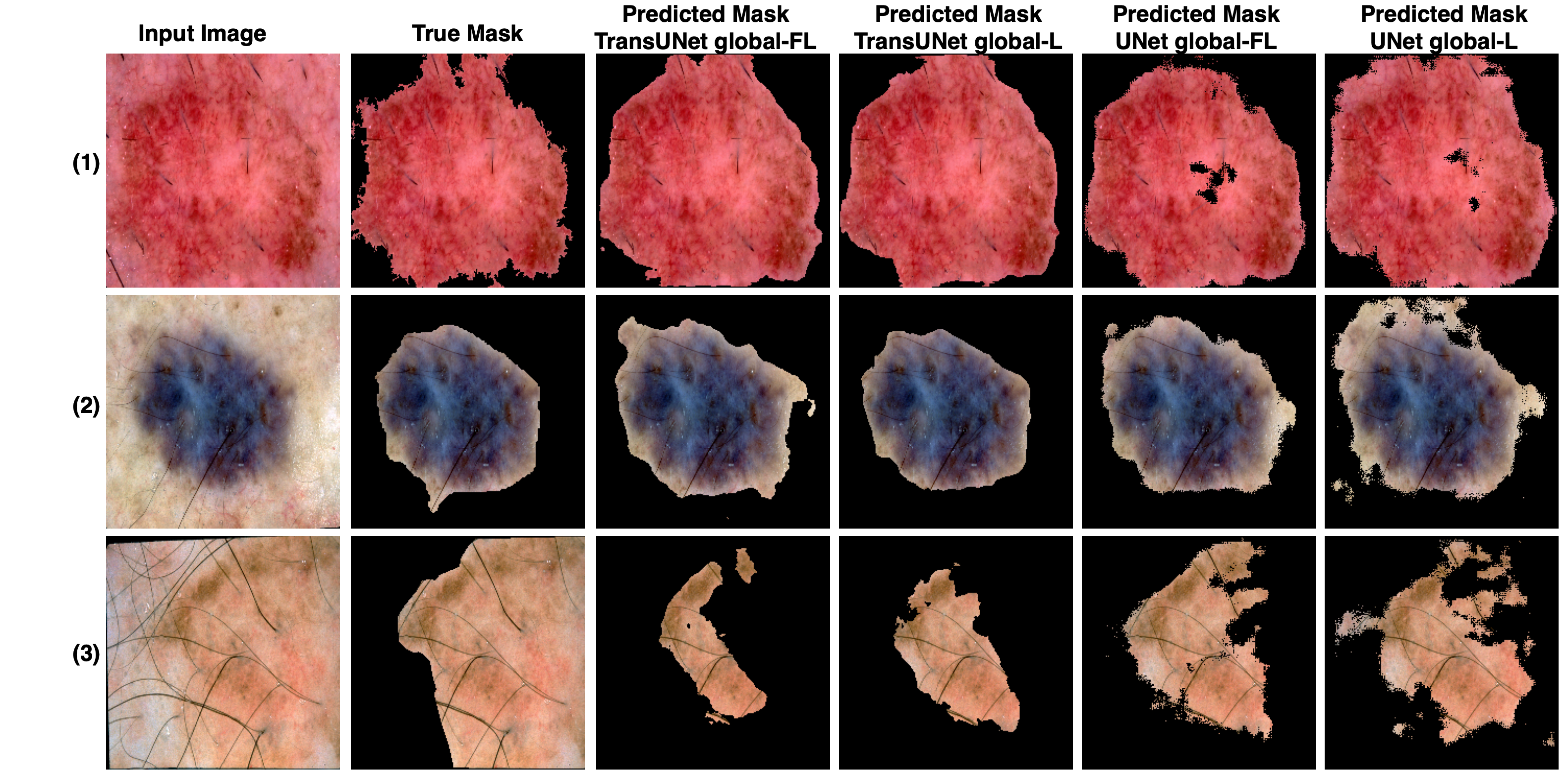}  % Adjust width as needed
    \caption{Pictorial representation of segmentation results generated by the  global-L and global-FL models for both TransUNet and UNet tested with ISIC 2018 dataset}
    \label{fig:qual:results}
\end{figure}
%%%%%%%%%%%%%%%%%%%%%%

Table \ref{tab:model_results} shows that for both UNet and TransUNet, the local models (local-1 to local-5) trained with local datasets perform inferior to the centrally trained global model (global-L), which has access to all client's images. This justifies that if a centralized model-building set-up exists, the centrally trained global model (global-L) would perform better than an individual client for the chosen dataset. The probable reason is that such a model would have access to more images, which generalized the input in a better way. The federated global model (global-FL) performs better than the best local model for both segmentation networks. This advocates that federated learning can capture the data diversity available to different clients without accessing the raw images. However, the performance of the federated global model (global-FL) is comparable to the centrally trained global model (global-L).

As per our visual qualitative segmentation performance evaluation, it is difficult to perceive the segmentation networks' performance gap between the federated global model (global-FL) and the centrally trained global model (global-L) for the majority of the images. For visual representation, we select three images and shows their expert annotated ground truth, and their segmentation outcome with the federated global model (global-FL) and global model (global-L) in Fig.~\ref{fig:qual:results}. It is evident that among these three images, the third image is not segmented well.

\begin{figure}[!b]
    \centering
    % First image at the left
    \includegraphics[width=0.995\textwidth]{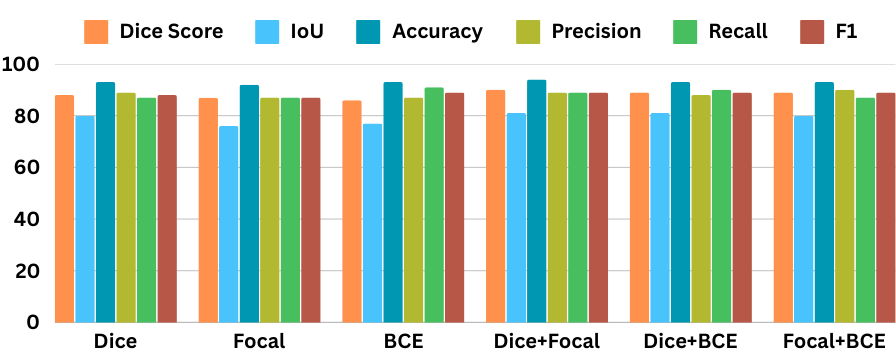}  % Adjust width as needed
    
    % Adding caption
    \caption{Results of the global-FL TransUNet model trained and tested on ISIC 2018 dataset considering Dice, Focal, BCE, Dice+Focal, Dice+BCE, and Focal+BCE losses}
    \label{fig:res:bar}
\end{figure}

Our second experiment validated the performances of the trained models obtained from the ISIC 2018 skin lesion segmentation challenge dataset with respect to $PH^2$ dataset (an external dataset). This will highlight the relative performance gap of the models and justify the importance of the federated global model. Table \ref{tab:TransUNet_results_PH2} contains the performance observed in $PH^2$ dataset. We find that the performances are degraded for all models trained with ISIC 2018 dataset. The probable reason is that the $PH^2$ dataset was prepared early, and the imaging device was not that much improved; thus, they might have more imaging artefacts, so segmentation is relatively difficult than the ISIC dataset. However, the relative performance gap among the models is quite similar to our first experiment. The global-FL model is performing much better than the best local model. The global-L and global-FL models for both UNet and TransUNet achieve very similar performances in terms of all of the performance metrics. Moreover, in some cases, global-FL beats global-L (for example, dice score in the case of UNet). In both our first and second experiments, TransUnet outperformed UNet by a significant margin, as shown in Tables \ref{tab:model_results} and \ref{tab:TransUNet_results_PH2}. For example, the Dice score for TransUNet is $\sim 3\%$ higher than the Dice for UNet (look at the rows for the global-L of TransUNet and the global-L of UNet in Table \ref{tab:TransUNet_results_PH2}). These results clearly show that TransUNet is more effective in segmenting skin lesions. Due to this, the rest of our experiments are only conducted for TransUNet.

\begin{figure}[!t]
    \centering
    % First image at the left
    \includegraphics[width=0.995\textwidth]{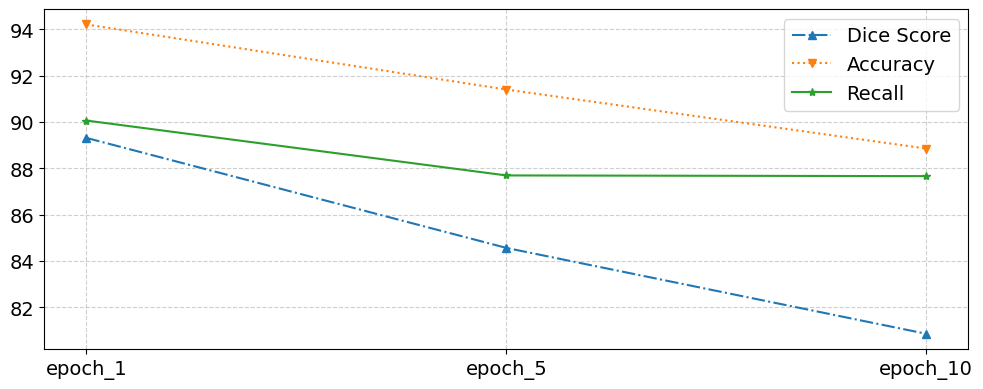}  % Adjust width as needed
    
    % Adding caption
    \caption{Performances yield by the global-FL TransUNet model trained and tested on the ISIC 2018 dataset, when the \#epochs in each iteration during federated learning is varied from 1 (epoch\_1) to 5 (epoch\_5) and 10 (epoch\_10).}
    \label{fig:res:loss}
\end{figure}

\justify
\textbf{Ablation Study.} The ablation study is conducted to choose the right loss function for training the networks. In this study, we train only the model global-FL to decide the loss function by selecting an optimal combination of different loss components. The loss components are the conventional segmentation losses: Dice, Focal, and BCE. Thus, the global-FL is trained for each of the following (joint) losses: Dice, Focal, BCE, Dice $+$ Focal, Dice $+$ BCE, and Focal $+$ BCE. All the models are trained with the training set of the ISIC 2018 skin lesion segmentation dataset, followed by the testing with its test set. Considering the various performance metrics, the results of this experiment are presented in a bar plot in Fig. \ref{fig:res:bar}. It can be seen that Dice $+$ Focal loss shows superior or similar performances compared to other losses. However, Dice $+$ Focal loss yields $\sim 1\%$ higher accuracy than others. Due to this, the combination of Dice and Focal losses is chosen as the loss function in our implementation.
\justify
\textbf{Selecting Number of Epochs ($E$) in An Iteration of Federated Learning.} We also experiment to determine an optimal number of epochs in an iteration of the federated learning to train the global model (global-FL) of TransUNet. The number of epochs per iteration is set to 1 (see epoch\_1), 5 (epoch\_5), and 10 (epoch\_10), and evaluated the performances of the global-FL TransUNet model. The Dice, accuracy, and F1 scores are shown in Fig. \ref{fig:res:loss}, where the model is trained and tested on the ISIC 2018 dataset. The best performance is seen when the number of epochs is set to one per iteration. Although setting one epoch per iteration is computationally expensive, we set this value in our implementation to obtain higher performances.

\section{Conclusion and future scope of Work}\label{sec:Conclusion}
This paper experimented with the federated learning algorithm for dermoscopic image segmentation. Two commonly known deep segmentation networks are trained with the federated learning algorithm in a simulated environment. ISIC 2018 skin lesion segmentation challenge dataset is used for training the model and testing. The performance is also analyzed with respect to the $PH^2$ dermoscopic image dataset. We find that the TransUNet is more suitable than UNet for skin lesion segmentation. The experimental outcome advocates that the global model trained with federated learning improves the segmentation performance in data restriction scenarios. However, in most cases, we find some performance gaps when a model is trained in a data restriction scenario rather than having full data access for centralized training. The performance obtained from the centrally trained global model shows that there is a huge scope to improve the segmentation performance so that the global model obtained from the federated learning will perform human-like. 

The imaging artifact, the presence of hair in the skin, creates dermatological images that are noisy and can mislead the training. Producing a reliable, trained model with such noisy images will be the immediate future work of this research. Developing a better deep network and utilising visual foundation models for skin lesion segmentation is important for future work. Moreover, the development of a federated learning algorithm for segmenting skin lesion attributes like pigmented networks, globules, etc., is to be conducted next. In future work, we will evaluate model performance under more nuanced non-IID conditions by employing advanced data partitioning strategies, such as the Dirichlet distribution-based sampling protocol. Enhancing security during model aggregation with privacy-preserving techniques like Differential Privacy, Homomorphoic Encryption is another important future scope of work.

\subsubsection*{Acknowledgements}
This work is partially supported by the Institute Research Initiation Grant (RIG) of the Indian Institute of Technology Jodhpur. We thank Rahul Kumar and Aayushman for helping us organize the data and implement the proposed technique.

\bibliographystyle{splncs04}
\bibliography{refs}

\end{document}